\pdfoutput=1
\documentclass[11pt]{article}
\usepackage[final]{acl}
\usepackage{times}
\usepackage{latexsym}
\usepackage[T1]{fontenc}
\usepackage[utf8]{inputenc}
\usepackage{microtype}
\usepackage{inconsolata}
\usepackage{graphicx}
\usepackage{booktabs}
\usepackage{multirow}
\usepackage{amsmath}
\usepackage{amssymb}    
\usepackage{pifont}     
\usepackage{xcolor} 
\newcommand{\checkmarkgreen}{\textcolor{green!60!black}{\ding{52}}}
\newcommand{\xmarkred}{\textcolor{red}{\ding{55}}}
\usepackage{gensymb} 

\makeatletter
\let\cas@beginabstract\abstract      
\let\cas@endabstract  \endabstract   
\makeatother

\usepackage{arabtex}
\usepackage{utf8}

\makeatletter
\let\abstract   \cas@beginabstract   
\let\endabstract\cas@endabstract     
\makeatother

\usepackage[utf8]{inputenc}
\usepackage{array}    
\usepackage{tabularx} 
\usepackage{url}      

\newcommand{\featurecheck}{\textcolor{green!60!black}{\ding{52}}} 
\newcommand{\featurecross}{\textcolor{red}{\ding{55}}}           
\newcommand{\featurepartial}{\textcolor{orange!80!black}{\textbf{Ltd.}}} 

\usepackage[T1]{fontenc}

\title{QARI-OCR: High-Fidelity Arabic Text Recognition through Multimodal Large Language Model Adaptation}

\author{
\textnormal{Ahmed Wasfy\textsuperscript{1,2\dag}} \quad
\textnormal{Omer Nacar\textsuperscript{3\dag}} \quad
\textnormal{Abdelakreem Elkhateb\textsuperscript{1}} \quad
\textnormal{Mahmoud Reda\textsuperscript{1}} \\
\textnormal{Omar Elshehy\textsuperscript{1}} \quad
\textnormal{Adel Ammar\textsuperscript{3}} \quad
\textnormal{Wadii Boulila\textsuperscript{3}} \\
\textsuperscript{1}NAMAA \\
\textsuperscript{2}KAND CA Corp. \\
\textsuperscript{3}Prince Sultan University \\
Emails: \texttt{onajar@psu.edu.sa}, \texttt{ahmed.wasfy@kand.ca} \\
\textsuperscript{\dag}Corresponding authors
}

\begin{document}
\maketitle
\setcode{utf8}

\begin{abstract}
The inherent complexities of Arabic script—its cursive nature, diacritical marks (tashkeel), and varied typography—pose persistent challenges for Optical Character Recognition (OCR). We present Qari-OCR, a series of vision-language models derived from Qwen2-VL-2B-Instruct, progressively optimized for Arabic through iterative fine-tuning on specialized synthetic datasets. Our leading model, QARI v0.2, establishes a new open-source state-of-the-art with a Word Error Rate (WER) of 0.160, Character Error Rate (CER) of 0.061, and BLEU score of 0.737 on diacritically-rich texts. Qari-OCR demonstrates superior handling of tashkeel, diverse fonts, and document layouts, alongside impressive performance on low-resolution images. Further explorations (QARI v0.3) showcase strong potential for structural document understanding and handwritten text. This work delivers a marked improvement in Arabic OCR accuracy and efficiency, with all models and datasets released to foster further research.
\end{abstract}
\section{Introduction}
Digital text accessibility is central to information preservation, dissemination, and analysis in today's data-driven world. While Optical Character Recognition (OCR) technology has made significant progress for Latin-based scripts, complex writing systems like Arabic continue to present substantial challenges. Arabic script, with its cursive nature, contextual character forms, diverse diacritical marks (tashkeel), and varied typographic styles, poses unique difficulties for conventional OCR approaches \cite{al2020review}.

Arabic is spoken by over 420 million people worldwide, making accurate Arabic OCR vital for cultural preservation, scholarly research, and information access~\cite{unesco2024arabic}. Despite this importance, existing Arabic OCR solutions often underperform compared to their Latin-script counterparts, with particularly poor handling of diacritical marks that significantly affect pronunciation and meaning.~\cite{alwajih2024peacock}

This paper introduces \textit{Qari-OCR}, a fine-tuned vision-language model based on \textit{Qwen2-VL-2B-Instruct}, specifically optimized for Arabic text recognition. \textit{Qari-OCR} was developed through an iterative process, with each version leveraging progressively enhanced synthetic datasets to improve performance on specific aspects of Arabic text, detailed in Table~\ref{tab:dataset_versions_optimized}. Our approach utilizes recent multimodal learning advances for superior Arabic OCR performance with computational efficiency.

\begin{table*}[htbp!]
\centering
\caption{Key Characteristics and Objectives of Qari-OCR Models Versions.} 
\label{tab:dataset_versions_optimized}
\footnotesize

\setlength{\aboverulesep}{0.3ex} 
\setlength{\belowrulesep}{0.3ex}

\begin{tabularx}{\textwidth}{
    >{\raggedright\arraybackslash}p{0.13\textwidth} 
    >{\raggedright\arraybackslash}X 
    >{\raggedright\arraybackslash}X 
    >{\centering\arraybackslash}p{0.08\textwidth} 
    >{\centering\arraybackslash}p{0.06\textwidth} 
    >{\centering\arraybackslash}p{0.06\textwidth} 
    >{\centering\arraybackslash}p{0.08\textwidth} 
    >{\centering\arraybackslash}p{0.10\textwidth} 
}
\toprule
\textbf{Model Ver.} & 
\textbf{Key Features/Focus} & 
\textbf{Objective/Tested Capability} & 
\textbf{Training Dataset Size} & 
\textbf{HTML?} & 
\textbf{Diacritics?} & 
\textbf{Layout Complexity?} & 
\textbf{Handwritten Support?} \\ 
\midrule
Qari-OCR v0.1 & Clean, no diacritics, 5 fonts, uniform min. size/layout. & Baseline on legible, low-noise data. & 5,000 & \xmarkred & \xmarkred & \xmarkred & \xmarkred \\
Qari-OCR v0.2 & Diacritics, broader typography (10 fonts), linguistic complexity. & Recognition of diacritic-rich/classical text. & 50,000 & \xmarkred & \checkmarkgreen & \xmarkred & \xmarkred \\
Qari-OCR v0.3 & Multi-font sizes/page (headers, body), realistic layouts. & Spatial parsing for mixed-size, complex layouts. & 10,000 & \checkmarkgreen & \checkmarkgreen & \checkmarkgreen & \checkmarkgreen \\
\bottomrule
\end{tabularx}
\end{table*}

Our key contributions include:
\begin{description}
    \item[State-of-the-Art Model:] We trained three different QARI-OCR Models,that significantly outperforms leading open-source solutions for full-page text recognition and different layout complexities.
    \item[Comprehensive Diacritical, and Script Support:] Qari-OCR models demonstrate comprehensive support for Arabic diacritical marks (tashkeel), including fathah, kasrah, dammah, sukun, shadda, and tanwin forms.
    \item[Evaluation \& Release:] We publicly release all trained models alongside their corresponding evaluation datasets and a standardized evaluation framework to enable reproducible research and facilitate downstream applications. For review, see \href{https://huggingface.co/collections/NAMAA-Space/qari-ocr-a-high-accuracy-model-for-arabic-optical-character-67c6cdff9584ef0684391335}{huggingface repository}.

\end{description}

The remainder of this paper is organized as follows: Section~\ref{sec:related_work} reviews related work in Arabic OCR. Section~\ref{sec:methodology} details our dataset generation pipeline and model training. Section~\ref{sec:experiments} presents the experimental setup and results. Section~\ref{sec:limitations} outlines the limitations of our models, and Section~\ref{sec:conclusion} concludes the paper and suggests future work.

\section{Related Work}
\label{sec:related_work}

\begin{table*}[t] 
\centering
\caption{Evolution of OCR Approaches and Key Characteristics Relevant to Arabic.}
\label{tab:related_work_ocr_summary} 
\footnotesize 
\setlength{\heavyrulewidth}{0.08em} 
\setlength{\lightrulewidth}{0.05em} 
\setlength{\cmidrulewidth}{0.03em} 
\setlength{\aboverulesep}{0.4ex}    
\setlength{\belowrulesep}{0.4ex}    

\begin{tabularx}{\textwidth}{%
    >{\raggedright\arraybackslash}X
    >{\centering\arraybackslash}p{2.0cm} 
    >{\centering\arraybackslash}p{2.2cm} 
    >{\centering\arraybackslash}p{2.2cm} 
    >{\centering\arraybackslash}p{1.8cm} 
    >{\raggedright\arraybackslash}X
}
\toprule
\textbf{OCR Approach Category} &
\textbf{End-to-End Learning} &
\textbf{Arabic Diacritic Handling} &
\textbf{Arabic Font/Style Diversity} &
\textbf{Multimodal Foundation} &
\textbf{Primary Focus / Example} \\
\midrule
Traditional OCR & \featurecross & \featurecross & \featurecross & \featurecross & Segmented classification; Early systems. \\ 
Early Deep Learning OCR (CNN-RNN-CTC) & \featurecheck & \featurepartial & \featurepartial & \featurecross & General text OCR; CRNN \cite{puigcerver2017multidimensional}. \\
Transformer-based OCR & \featurecheck & \featurepartial & \featurepartial & \featurecross & Sequence modeling for text; TROCR \cite{li2023trocr}. \\
Early Arabic-Specific DL OCR & \featurecheck & \featurepartial & \featurecheck & \featurecross & Targeted Arabic data; \cite{yousef2020accurate}. \\
Specialized Arabic Foundation Models (e.g., Qalam) &
\featurecheck &
\featurecheck &
\featurecheck &
\featurecheck &
Deep Arabic script recognition; Qalam \cite{bhatia2024qalam}. \\
General MLLMs (e.g., Qwen2-VL) & \featurecheck & \featurepartial & \featurepartial & \featurecheck & Broad vision-lang. tasks; Qwen2-VL \cite{wang2024qwen2}. \\ 
Arabic-Inclusive MLLMs (e.g., AIN) & \featurecheck & \featurecheck & \featurecheck & \featurecheck & Broader Arabic multimodal; AIN \cite{heakl2025ain}. \\ 
\textbf{Qari-OCR (Our Work)} & \textbf{\featurecheck} & \textbf{\featurecheck} & \textbf{\featurecheck} & \textbf{\featurecheck} & \textbf{Specialized Arabic OCR via MLLM.} \\
\bottomrule
\end{tabularx}
\end{table*}

The journey of Optical Character Recognition (OCR) has been marked by a significant evolution from early, rule-heavy systems to sophisticated deep learning paradigms, each step bringing new capabilities and addressing longstanding challenges, particularly for complex scripts like Arabic. Initial OCR approaches often involved a structured pipeline of preprocessing, explicit character segmentation, and traditional classification techniques \cite{jasim2020arabic,graves2008offline}. However, these methods encountered substantial difficulties with the cursive, context-sensitive nature of Arabic script \cite{alrobah2022arabic}. 

The advent of deep learning, particularly Convolutional Neural Networks (CNNs) combined with Recurrent Neural Networks (RNNs), as exemplified by the CRNN model \cite{puigcerver2017multidimensional}, revolutionized the field. Such architectures enabled end-to-end learning, implicitly handling segmentation and significantly improving performance on general text. Transformer models later pushed the boundaries further with their powerful attention mechanisms, enhancing sequence modeling for OCR tasks, leading to models like TROCR \cite{li2023trocr}.

Developing effective OCR for Arabic necessitates a keen focus on its unique orthographic characteristics: right-to-left text flow, intricate ligatures, contextually varying character shapes, and the varied positional placements of diacritical marks in Arabic. While deep learning advancements provided a robust foundation, specialized efforts were made to adapt these for Arabic. This included training on Arabic-specific datasets and tailoring architectures, as seen in some deep learning applications for Arabic OCR \cite{yousef2020accurate}. More recently, dedicated models like Qalam \cite{bhatia2024qalam}, a multimodal system built on a Swinv0.2 encoder and RoBERTa decoder, have been developed specifically for Arabic OCR and handwriting recognition. 

The latest wave of innovation involves Multimodal Large Language Models (MLLMs), which aim to unify vision and language understanding for a wide array of tasks. These models can perform OCR as one of their many capabilities. The Qwen2-VL series \cite{wang2024qwen2} represents a significant advancement in general-purpose MLLMs, incorporating features like dynamic image resolution processing and effective multimodal fusion, leading to competitive performance on broad multimodal benchmarks. However, the inherent design of such generalist MLLMs does not typically optimize them for the high-fidelity, nuanced requirements of Arabic OCR. While MLLMs like AIN \cite{heakl2025ain} are emerging, trained with substantial authentic Arabic data to address various multimodal tasks including OCR, the specific challenges of detailed text recognition in Arabic images, especially concerning diacritics and diverse typography, continue to be an area noted for further improvement.

Our work, \textit{Qari-OCR}, is positioned within this evolving landscape. We leverage the sophisticated architecture of a general-purpose MLLM, specifically \textit{Qwen2-VL-2B-Instruct} \cite{wang2024qwen2}, as a foundational model. Through specialized fine-tuning on our synthetic Arabic text datasets, we adapt this MLLM to function effectively as an OCR system. By combining targeted dataset curation and parameter-efficient adaptation, Qari-OCR addresses accurate diacritic recognition and varied font styles, advancing high-fidelity Arabic text recognition. The evolution and comparative characteristics of these OCR approaches are summarized in Table~\ref{tab:related_work_ocr_summary}.

\begin{figure*}[t]
 \centering
  \includegraphics[width=0.7\textwidth]{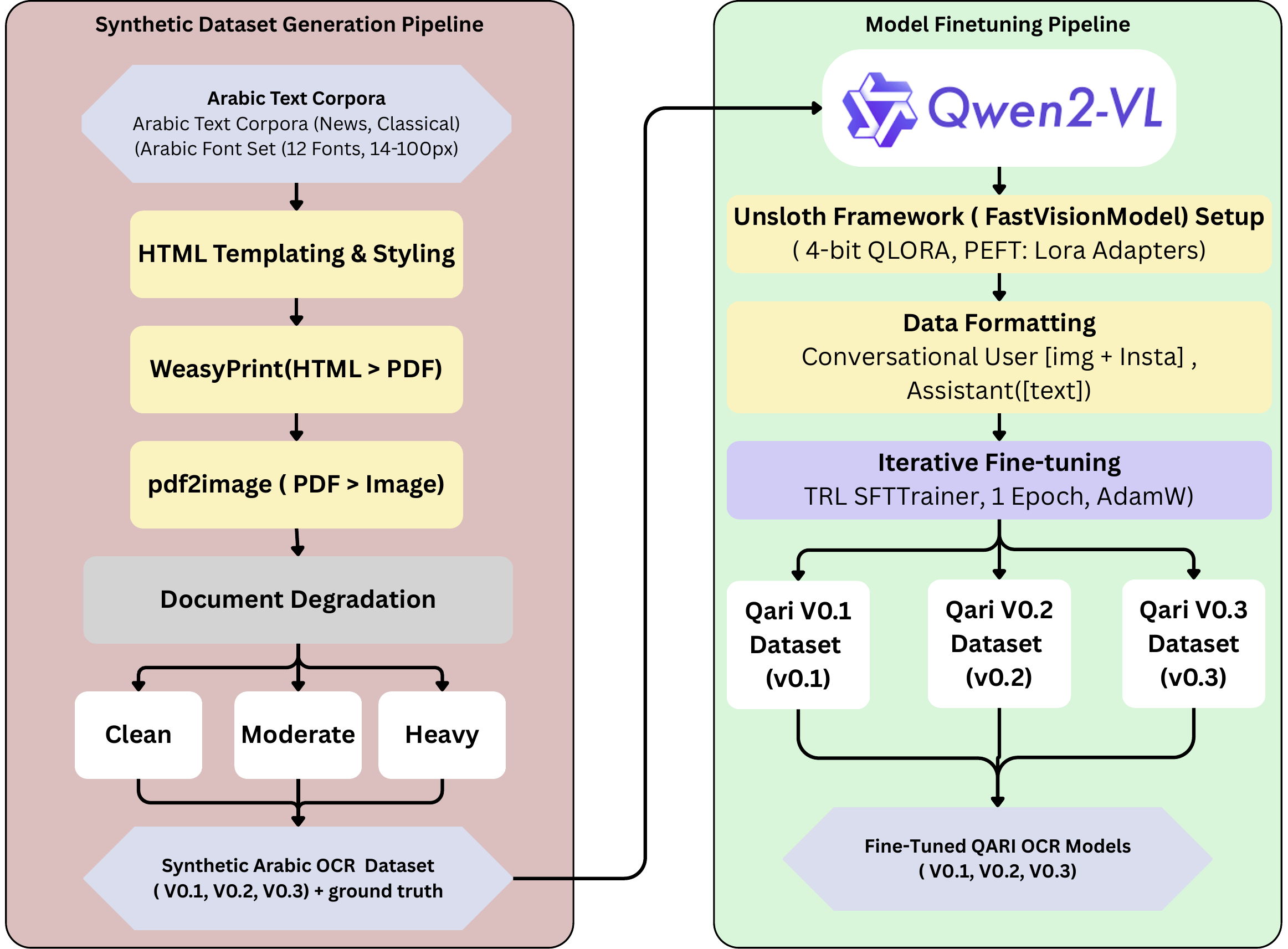}
  \caption{Qari-OCR Dataset Generation and Model Training Pipeline}
  \label{fig:methodology_workflow}
\end{figure*}
\section{Methodology}
\label{sec:methodology}
The development of Qari-OCR was implemented through a two-stage methodological framework: firstly, the generation of diverse synthetic datasets engineered to encapsulate the complexities of Arabic script; and secondly, the iterative fine-tuning of an advanced vision-language model using these specialized datasets. An illustrative overview of this workflow is presented in Figure~\ref{fig:methodology_workflow}.

\subsection{Synthetic Dataset Generation for QARI}

\label{ssec:dataset_generation_summary}

To bridge gaps in existing Arabic OCR corpora—namely diacritic coverage, font diversity, and realistic layouts—we devised a three-stage synthetic data pipeline. Two complementary text sources were used: a modern news article collection and a classical Islamic corpus (rich in tashkīl). The text was rendered programmatically in HTML using twelve distinct Arabic fonts (from common Naskh to ornate calligraphic styles) at sizes varying between 14 px and 100 px, then converted to PDF via \textit{WeasyPrint}~\footnote{https://weasyprint.org} and to images via \textit{pdf2image}~\footnote{https://pdf2image.readthedocs.io/en/latest/index.html}.

\begin{itemize}

\item \textbf{Dataset v0.1:} Non-diacritized text, a limited font set, and uniform minimal size establish a high-legibility baseline.

\item \textbf {Dataset v0.2:} The dataset v0.2 introduces full diacritics and expands the font repertoire to enhance the recognition of vocalized and classical texts.

\item \textbf{Dataset v0.3:} Introduces mixed font sizes on each page to simulate realistic document structures (headers, body, annotations) and HTML spatial/layout parsing.

\end{itemize}

Finally, each image undergoes one of three synthetic degradation treatments—\emph{Clean}, \emph{Moderately Degraded} (subtle noise, color shifts, mild blur), or \emph{Heavily Degraded} (textured backgrounds, aggressive blur)—with all variants paired to their ground‐truth transcription. This progression yields a robust, multi-faceted Arabic OCR dataset suitable for training and evaluating QARI across increasing levels of linguistic, typographic, and visual complexity.

\subsection{Model Architecture and Training Strategy}

\label{ssec:model_training_summary}

We built Qari-OCR on the \textit{Qwen2-VL-2B-Instruct} backbone \cite{wang2024qwen2}, leveraging its Naive Dynamic Resolution for adaptive image scaling and M-RoPE for robust cross-modal positional embeddings. To optimize fine-tuning efficiency, we optionally quantized the model to 4-bit and inserted LoRA adapters (rank = 16) into both vision and language modules.

Training data comprised conversationally formatted image–text pairs, where each “user” message carried an image and prompt, and the “assistant” reply provided the ground-truth Arabic transcription. We conducted three matched fine-tuning runs, each on a different synthetic dataset version, as summarized in~\ref{tab:dataset_versions_optimized}.

All models were fine-tuned for a single epoch using the Unsloth library. footnote{\url{https://github.com/unslothai/unsloth}} with the AdamW optimizer~\cite{loshchilov2017decoupled} and with $learning\_rate$ equal to 2e-4 and $weight\_decay$ of 0.01 with linear $lr\_scheduler$. Input images were resized and normalized to Qwen2-VL specifications, and training was orchestrated with Hugging Face’s\texttt{SFTTrainer}\footnote{\url{https://huggingface.co/docs/trl/en/sft_trainer}} using the \texttt{UnslothVisionDataCollator}, a per-device batch size of 2, and 4 gradient-accumulation steps (effective batch size = 8). All experiments ran on a single NVIDIA A6000 GPU (48 GB VRAM).

\begin{figure*}[t]
    \centering
    \includegraphics[width=\textwidth]{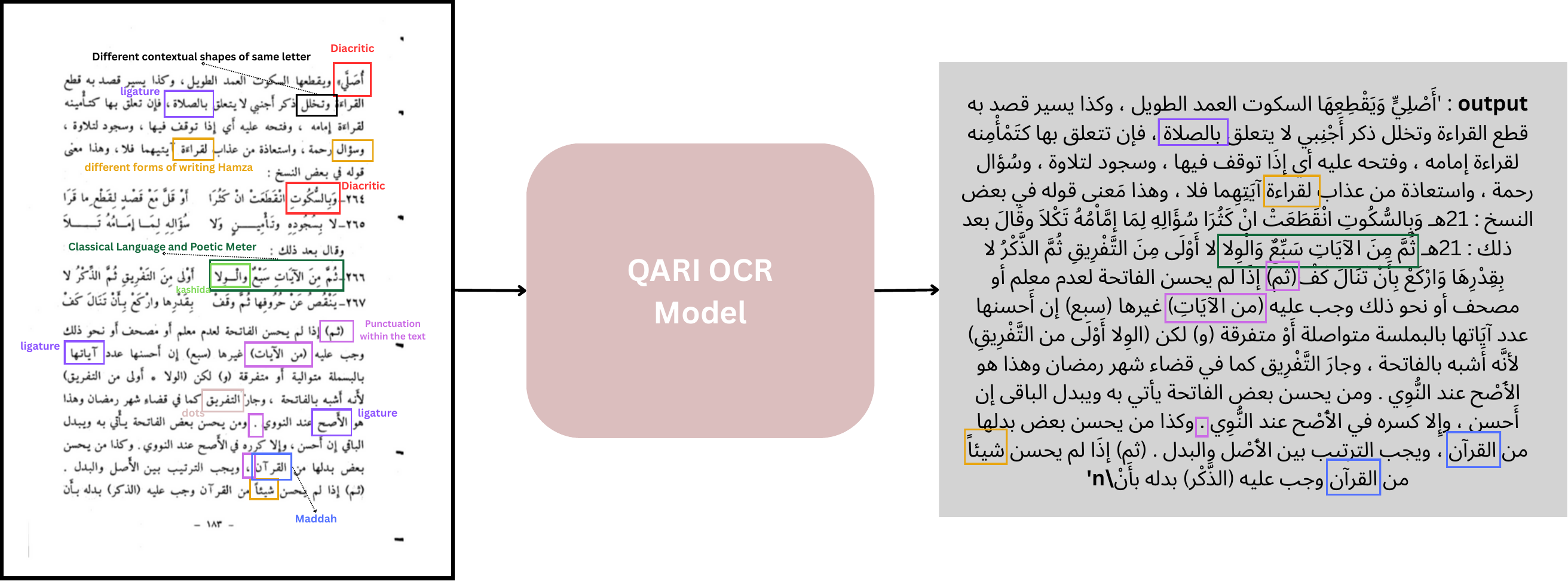}
    \caption{Qualitative example demonstrating Qari-OCR's handling of various Arabic script complexities. The input image (left, with annotations highlighting features like diacritics, ligatures, contextual shapes, etc.) is processed by the Qari-OCR model, producing the transcribed text output (right).}
    \label{fig:qari_ocr_challenges_section} 
\end{figure*}

\section{Experimental Results}
\label{sec:experiments}

This section presents the experimental setup, evaluation metrics, and empirical results used to benchmark Qari-OCR and selected baselines on challenging Arabic text.

\subsection{Experimental Setup}
\label{ssec:exp_setup}
We assembled a test set of 200 scanned pages from traditional Arabic print—complete with diacritics, complex ligatures, and dense layouts—to mirror the challenges of historical and scholarly texts. Every image underwent the same generic preprocessing (no language-specific tweaks or manual annotations), ensuring a fair evaluation of each OCR system’s raw performance.

Our benchmark suite comprises Qari-OCR and six different OCR systems spanning from classical engines to cutting-edge vision–language models: Tesseract OCR~\cite{smith2007overview}, EasyOCR~\cite{pattanayak2023novel}, Mistral OCR~\cite{mistralOCR2025}, AIN~\cite{heakl2025ain}, Qwen 2.5-7B Instruct~\cite{wang2024qwen2}, and Qwen 2-7B~\cite{wang2024qwen2}.

\begin{table}[h] 
\centering
\caption{Comparative Performance of OCR Models on the Arabic Test Set. Lower CER/WER and higher BLEU indicate better performance.} 
\label{tab:main_ocr_results_compact} 
\small 
\setlength{\tabcolsep}{4pt} 

\begin{tabular}{@{} l rrr @{}} 
\toprule
\textbf{Model} & \textbf{CER} $\downarrow$ & \textbf{WER} $\downarrow$ & \textbf{BLEU} $\uparrow$ \\
\midrule
Tesseract OCR & 0.436 & 0.889 & 0.108 \\
EasyOCR & 0.791 & 0.918 & 0.051 \\
Mistral OCR (API-based) & 0.210 & 0.440 & 0.570 \\
AIN & 0.640 & 0.830 & 0.210 \\
Qwen 2.5-7B Instruct & 0.550 & 0.800 & 0.220 \\
Qwen 2-7B & 0.740 & 1.050 & 0.160 \\
\midrule
QARI v0.1 (Ours) & 1.915 & 2.025 & 0.221 \\
\textbf{QARI v0.2 (Ours)} & \textbf{0.061} & \textbf{0.160} & \textbf{0.737} \\
QARI v0.3 (Ours) & 0.300 & 0.485 & 0.545 \\
\bottomrule
\end{tabular}
\end{table}

\subsection{Evaluation Metrics}
\label{ssec:evaluation_metrics}
To assess OCR performance on Arabic text, we employed three complementary metrics: Character Error Rate (CER), Word Error Rate (WER)~\cite{klakow2002testing}, and the BLEU score~\cite{papineni2002bleu}. These metrics provide a holistic view, capturing errors from fine-grained character inaccuracies to broader linguistic and structural deviations.

CER measures the normalized Levenshtein distance at the character level between the predicted and ground-truth text. It is particularly critical for Arabic OCR due to its sensitivity to errors in diacritics and morphologically complex character sequences, which significantly impact meaning. WER, operating similarly at the word level, reflects how recognition errors affect sentence structure and is useful for identifying tokenization or word-level mistakes common in processing Arabic script. Finally, the BLEU score, by assessing n-gram overlap, offers insights into phrase-level fidelity and the preservation of fluent text structure, which is valuable for evaluating the overall readability and coherence of the OCR output.

\subsection{Results}
\label{ssec:results_analysis} 
The comparative performance of our Qari-OCR model versions (QARI v0.1, 0.2, and v0.3) and the selected models was assessed using our Arabic test set. The quantitative outcomes, based on CER, WER, and BLEU scores, are presented in Table~\ref{tab:main_ocr_results_compact}.

\begin{figure*}[t]
    \centering
    \includegraphics[width=1.9\columnwidth]{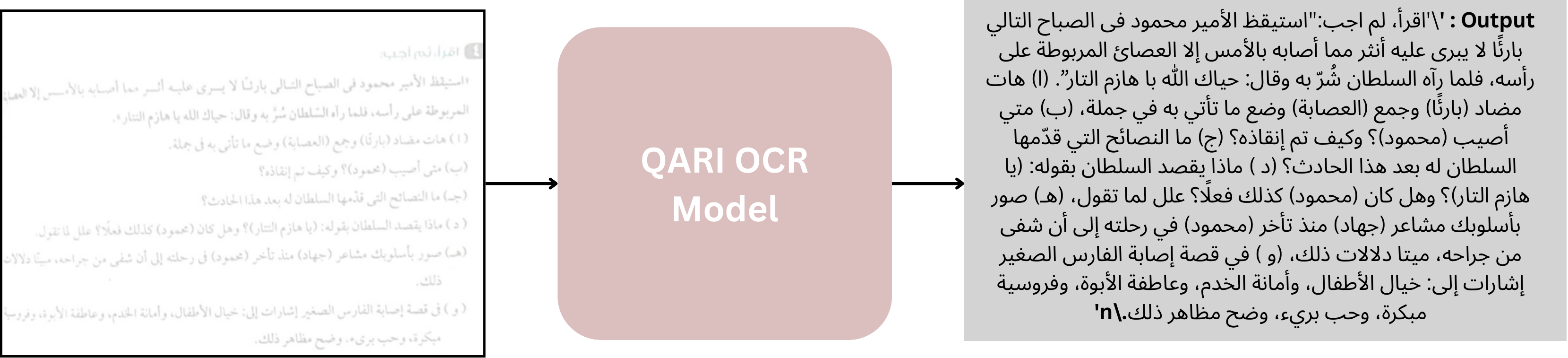}
    \caption{Example of Qari-OCR (v0.3) accurately transcribing Arabic text from a low-resolution and tightly cropped image, showcasing robustness to visual constraints.}
    \label{fig:low_resolution_ocr}
\end{figure*}

\begin{figure}[htbp!]
    \centering
    \includegraphics[width=0.9\columnwidth]{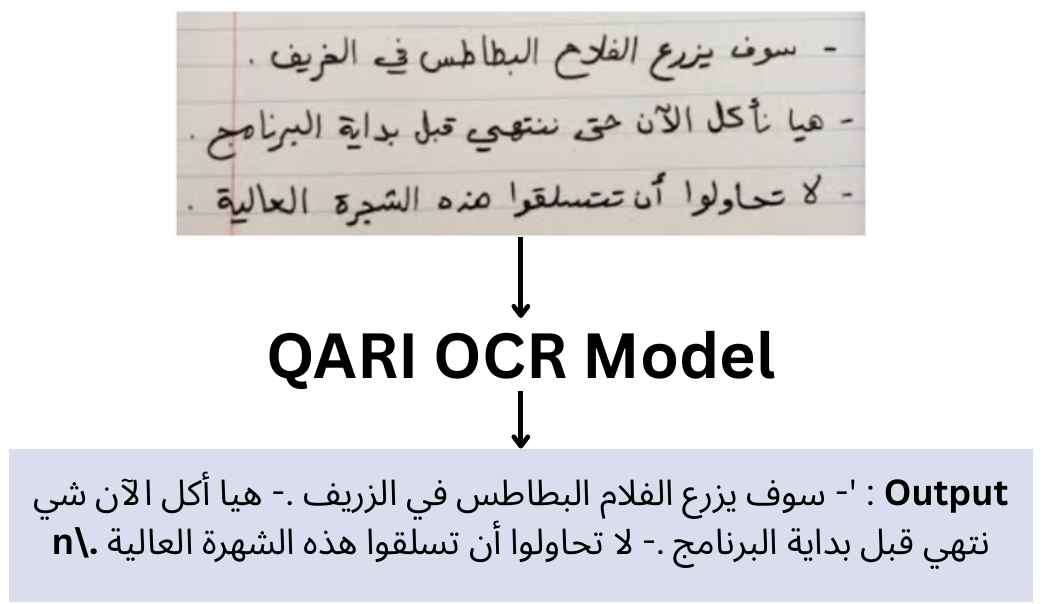}
    \caption{Qari-OCR v0.3 successfully transcribing handwritten Arabic text, maintaining sentence structure, punctuation, and recognizing itemized formatting.}
    \label{fig:handwritten_ocr}
\end{figure}

As shown in Table~\ref{tab:main_ocr_results_compact}, QARI v0.2 significantly outperforms all other open source models evaluated, establishing a new benchmark on our test set with a CER of 0.061, a WER of 0.160, and a BLEU score of 0.737. These results underscore the effectiveness of our specialized fine-tuning methodology, particularly the benefit derived from training on synthetic data rich in diacritics and typographic variations (Dataset v0.2). Notably, QARI v0.2 also surpasses the performance of the API-based Mistral OCR in terms of error rates and BLEU score. In contrast, the baseline Qwen models, without specific fine-tuning for Arabic OCR, demonstrate considerably higher error rates, emphasizing the critical need for task-specific adaptation when dealing with complex scripts.

In addition to these quantitative benchmarks, a qualitative assessment is vital for understanding the model's proficiency in handling the nuanced complexities of Arabic script. Figure~\ref{fig:qari_ocr_challenges_section} provides a visual illustration of Qari-OCR's output on a challenging text sample. The input image (left panel of Figure~\ref{fig:qari_ocr_challenges_section}) exhibits several features typical of printed Arabic that pose difficulties for OCR systems. These include the full array of diacritics (tashkeel) essential for pronunciation and meaning; ligatures such as Lam-Alif (\<لا>); contextually variant letterforms; classical language structures and poetic meter conventions; embedded punctuation and Eastern Arabic numerals; diverse orthographic forms of the Hamza (\<ء>); and features like Maddah (\<آ>) and crucial letter-distinguishing dots.

The corresponding output from our Qari-OCR model (right panel of Figure~\ref{fig:qari_ocr_challenges_section}) showcases a high degree of fidelity in transcribing these intricate elements. The model proficiently recognizes the majority of diacritical marks, accurately segments words despite ligatures and contextual letter shaping, and correctly renders classical linguistic forms. This qualitative performance provides strong corroborative evidence for the quantitative results, especially for QARI v0.2, highlighting its robustness in managing the various challenges frequently encountered in real‐world Arabic textual scripts.

Beyond quantitative benchmarks, qualitative analysis is crucial for understanding the model's practical capabilities. Figure~\ref{fig:qari_ocr_challenges_section} illustrates Qari-OCR's proficiency in handling different complexities, supporting the strong quantitative performance of QARI v0.2.

Furthermore, the model's resilience to optical degradation and its ability to handle varied inputs were tested. As shown in Figure~\ref{fig:low_resolution_ocr}, Qari-OCR (specifically QARI v0.3, trained on more complex layouts) accurately transcribes text from a low-resolution image. Despite the image’s small size and tightly cropped boundaries, the model robustly detects and transcribes the Arabic text, demonstrating its effectiveness with compressed layouts, edge-bound scripts, and reduced-resolution content. This capability is vital for digitizing real-world historical or educational Arabic materials, which may not always be of pristine quality.

In addition to printed text, QARI v0.3 was also assessed for its ability to process handwritten Arabic, a notoriously challenging task. Figure~\ref{fig:handwritten_ocr} illustrates its performance on a handwritten sample. The model accurately detects full sentences, preserving punctuation and word boundaries. Notably, it correctly interprets visual structural cues, such as itemized lists (akin to bullet points) and sentence-level formatting, even with the inherent variability of handwriting. This shows promising initial capabilities for handling handwritten Arabic content.

These qualitative examples, particularly from QARI v0.3 which was trained on more diverse layouts, complement the quantitative results and highlight the practical utility of Qari-OCR in handling a range of challenging real-world Arabic document types.

\begin{table*}[t]
\centering
\caption{CER, WER, and BLEU Score results by Font and Model on SARD Dataset}
\label{tab:sard_results}
\small
\begin{tabular}{@{}llccccc@{}}
\toprule
\textbf{Metric} & \textbf{Model} & \textbf{Amiri} & \textbf{Arial} & \textbf{Calibri} & \textbf{Sakkal M.} & \textbf{Scheherazade} \\
\midrule
\multirow{3}{*}{CER↓}
& Mistral OCR         & \textbf{0.011} & \textbf{0.051} & \textbf{0.035} & \textbf{0.040} & \textbf{0.020} \\
& Qari v0.2    & 0.200          & 0.230          & 0.193          & 0.216          & 0.156 \\
& Qari v0.3    & 0.350          & 0.461          & 0.400          & 0.424          & 0.483 \\
\midrule
\multirow{3}{*}{WER↓}
& Mistral OCR          & \textbf{0.041} & \textbf{0.248} & \textbf{0.166} & \textbf{0.194} & \textbf{0.099} \\
& Qari v0.2    & 0.267          & 0.308          & 0.249          & 0.293          & 0.211 \\
& Qari v0.3    & 0.369          & 0.482          & 0.432          & 0.449          & 0.464 \\
\midrule
\multirow{3}{*}{BLEU↑}
& Mistral OCR          & \textbf{0.920} & 0.634          & 0.746          & 0.715          & \textbf{0.845} \\
& Qari v0.2    & 0.723          & \textbf{0.703} & \textbf{0.745} & \textbf{0.701} & 0.782 \\
& Qari v0.3    & 0.346          & 0.229          & 0.286          & 0.279          & 0.255 \\
\bottomrule
\end{tabular}
\end{table*}

To evaluate robustness across diverse Arabic fonts, we benchmarked best-performing models, including QARI v0.2, QARI v0.3, and Mistral OCR, on the SARD dataset\footnote{https://huggingface.co/datasets/riotu-lab/SARD}, which includes 1,000 images spanning five common fonts, including Amiri, Arial, Calibri, Sakkal Majalla, and Scheherazade. 

As shown in Table~\ref{tab:sard_results}, Mistral achieved the lowest error rates overall, particularly excelling in CER and WER. However, QARI v0.2 was highly competitive—outperforming Mistral OCR in BLEU for the Arial font and matching it closely for Calibri. Notably, QARI v0.2's BLUE scores outperformed Mistral OCR for some fonts, including Arial, Calibri, and Sakkak, and consistently outperformed QARI v0.3 across all metrics. These results highlight QARI v0.2 as a strong open-source alternative, balancing accessibility, performance, and versatility across typographic variations.

\subsubsection{Impact of Model Quantization}

\label{sssec:quantization_impact}

To assess the trade-offs between model size, computational efficiency, and performance, we evaluated different quantization levels for our QARI v0.2 and QARI v0.3 models. Specifically, we compared versions fine-tuned or inferred using 8-bit precision against those utilizing more aggressive 4-bit quantization. The results, presented in Table~\ref{tab:quantization_results_compact}, highlight the impact of these quantization strategies on the CER, WER, and BLEU scores.

\begin{table}[htbp!] 
\centering
\caption{Performance of QARI with 8-bit Vs. 4-bit Quantization.}
\label{tab:quantization_results_compact}
\small 
\setlength{\tabcolsep}{4pt} 
\begin{tabular}{@{} l l rrr @{}} 
\toprule
\textbf{Model} & \textbf{Quant.} & \textbf{CER} $\downarrow$ & \textbf{WER} $\downarrow$ & \textbf{BLEU} $\uparrow$ \\
\midrule
QARI v0.2 & 8-bit & 0.091 & 0.255 & 0.583 \\
          & 4-bit & 3.452 & 4.516 & 0.001 \\
\addlinespace 
QARI v0.3 & 8-bit & 0.133 & 0.353 & 0.472 \\
          & 4-bit & 3.228 & 6.428 & 0.001 \\
\bottomrule
\end{tabular}
\end{table}

As observed in Table~\ref{tab:quantization_results_compact}, employing 8-bit quantization during fine-tuning or inference maintains strong performance for both QARI v0.2 and QARI v0.3, offering a good balance between efficiency and accuracy. However, the more aggressive 4-bit quantization leads to a substantial degradation in performance across all metrics for both model versions. This suggests that while 4-bit quantization significantly reduces the model footprint and can accelerate inference, it incurs a considerable accuracy cost for the fine-grained task of Arabic OCR with these specific models and fine-tuning parameters. The 8-bit versions, therefore, represent the more practical choice when accuracy is paramount, while 4-bit might be considered only in scenarios with extreme computational constraints where a significant drop in accuracy is acceptable.
\begin{figure*}[t]
    \centering
    \includegraphics[width=\textwidth]{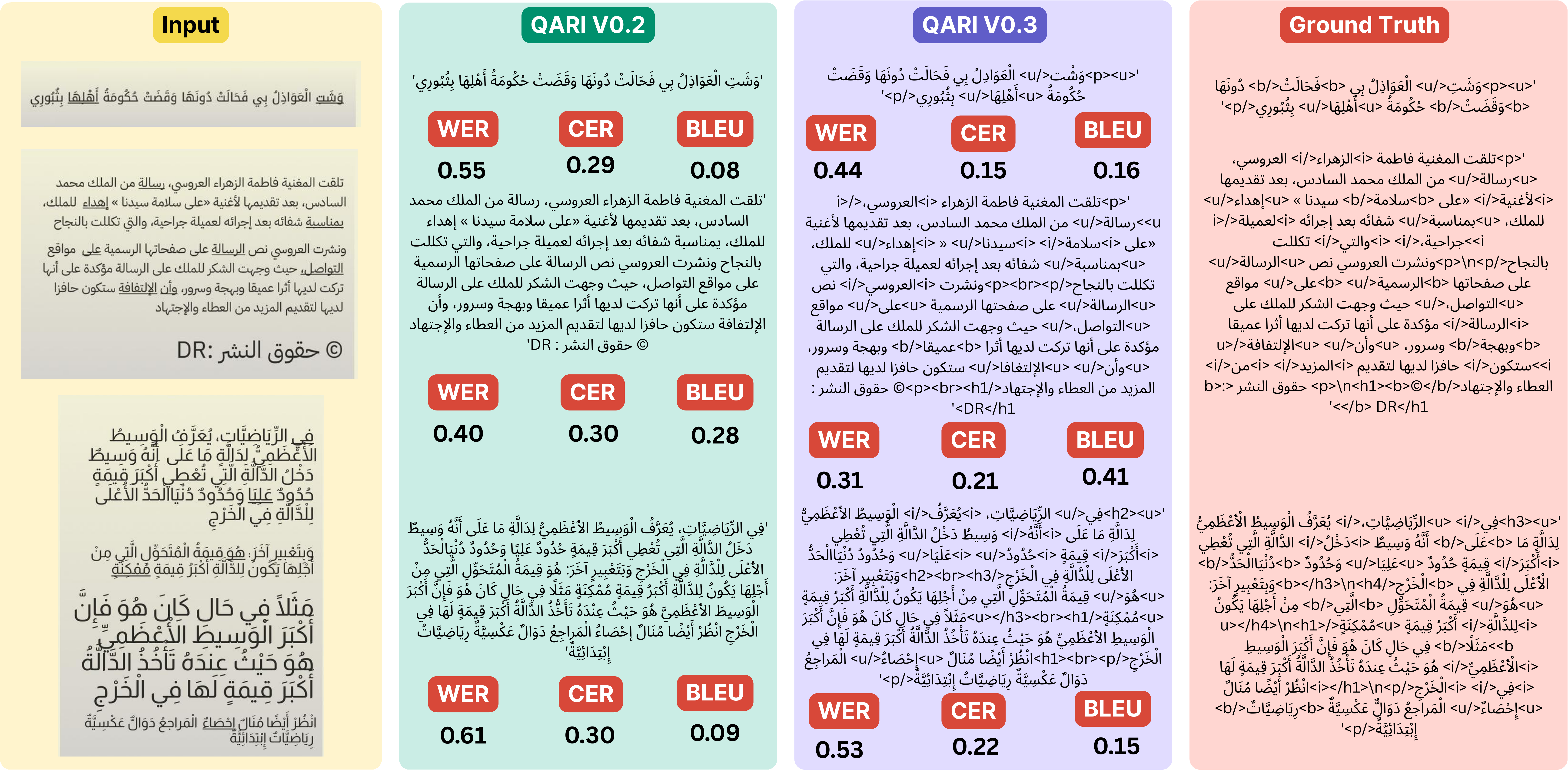} 
    \caption{Qualitative comparison of QARI v0.2 and QARI v0.3 outputs against Input and Ground Truth for various Arabic text samples.}
    \label{fig:qari_qualitative_comparison}
\end{figure*}

\begin{figure}[ht]
    \centering
    \includegraphics[width=0.7\columnwidth]{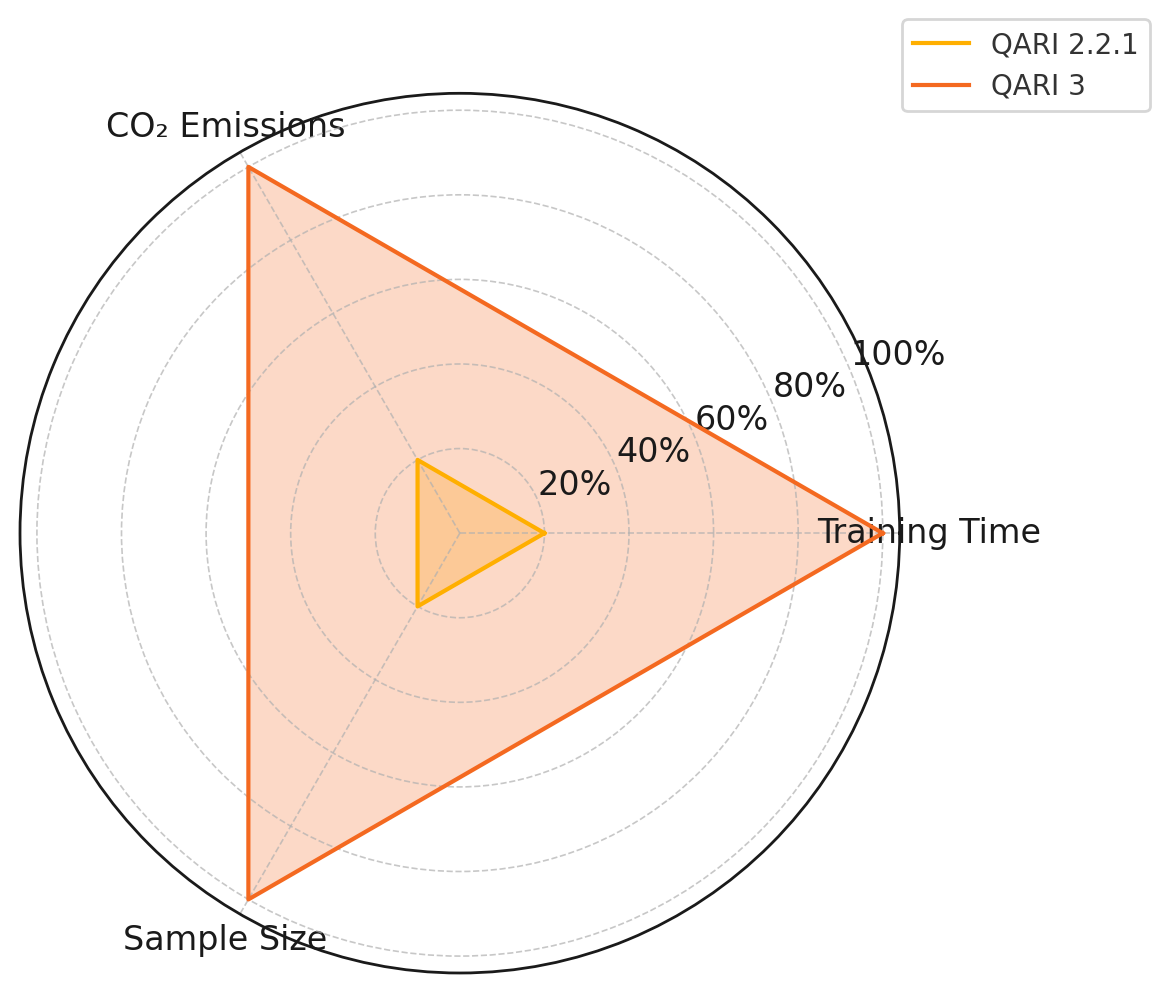} 
    \caption{Comparison of estimated resource consumption (CO2 Emissions, Training Time, Sample Size) for training QARI model variants.}
    \label{fig:resource_comparison_radar}
\end{figure}

\section{Discussion}
\label{sec:discussion}
Our experiments reveal distinct strengths across the Qari-OCR model iterations. While QARI v0.2, trained on 50,000 diverse samples (Dataset v0.2), demonstrates superior overall quantitative performance for plain text recognition (Table~\ref{tab:main_ocr_results_compact}), QARI v0.3, developed with a smaller 10,000-sample dataset focused on complex HTML-like layouts (Dataset v0.3), excels in preserving document structure.

Qualitative analysis, as shown in Figure~\ref{fig:qari_qualitative_comparison}, illustrates that QARI v0.3 effectively reconstructs HTML tags and formatting from input images, often achieving lower local error rates on these structurally rich examples compared to QARI v0.2's plain text output. This proficiency stems directly from QARI v0.3's targeted training on layout-aware synthetic data. The trade-off appears to be that QARI v0.2's larger and more varied character-level training data fostered better general textual accuracy, whereas QARI v0.3's smaller, specialized dataset, combined with a single training epoch, prioritized structural fidelity, potentially at the cost of some raw text accuracy on average.

Furthermore, resource efficiency considerations favor the QARI v0.3 approach for structure-oriented tasks. As depicted in Figure~\ref{fig:resource_comparison_radar}, the 10k-sample training regimen (represented by QARI v0.3's development) was significantly more economical in terms of training time and estimated CO2 emissions (1.88 kg eq. CO2 over ~11 hours) compared to the 50k-sample training (represented by "QARI 3", akin to QARI v0.2's development, at 9.4 kg eq. CO2 over ~55 hours). This highlights the potential for developing specialized, efficient models when the primary objective is structural document conversion.

In essence, QARI v0.2 serves as our most robust general-purpose Arabic OCR engine for accurate plain text extraction. QARI v0.3, however, validates a promising and resource-efficient strategy for applications requiring the understanding and reproduction of document structure, like HTML. The optimal model choice is therefore contingent on the specific end-goal: high-fidelity plain text output (QARI v0.2) or structural document reconstruction with greater training efficiency.

\section{Limitations}
\label{sec:limitations}
Despite the strong performance of Qari-OCR, particularly QARI v0.2, the current study and model possess certain limitations; Firstly, while proficient with dense printed text, the model may encounter difficulties with extremely heavy text layouts where character or line spacing is minimal, potentially leading to recognition errors. Secondly, Qari-OCR's current capabilities are primarily focused on textual content within the main body of documents; it often struggles to accurately recognize and extract text embedded within figures, charts, or complex graphical elements. Thirdly, the model's performance on historical or non-standard Arabic numeral systems has not been extensively validated and may be suboptimal. Finally, text elements typically found on the periphery of scanned pages, such as book titles on covers, page numbers, or marginalia, are sometimes skipped or inaccurately transcribed, indicating an area for improved contextual awareness and layout analysis.

\section{Conclusion}
\label{sec:conclusion}
In conclusion, this paper presented Qari-OCR, a fine-tuned vision-language model that achieves state-of-the-art performance for Arabic text recognition by leveraging extensive synthetic data and specializing the Qwen2-VL architecture. Our QARI v0.2 model significantly surpasses existing open-source solutions in accurately handling diacritics, diverse fonts, and complex layouts in printed Arabic. Future work will focus on addressing current limitations by enhancing robustness to dense text and embedded graphics, improving numeral recognition, advancing layout analysis for peripheral text, and extending capabilities to Arabic handwriting recognition. These efforts aim to develop Qari-OCR into an even more comprehensive solution for Arabic document understanding.

\section*{Acknowledgments}

The authors thank Prince Sultan University for their support.


\bibliography{custom}

\end{document}